\documentclass[conference]{IEEEtran}
\IEEEoverridecommandlockouts
\usepackage{amsmath,amssymb,amsfonts}
\usepackage{algorithm}
\usepackage{algpseudocode}
\usepackage{graphicx}
\usepackage{booktabs}
\usepackage{url}
\usepackage{textcomp}
\usepackage{xcolor}
\usepackage[numbers]{natbib}
\usepackage{booktabs}
\usepackage{tikz}
\usetikzlibrary{arrows.meta, positioning, fit, backgrounds}

\def\BibTeX{{\rm B\kern-.05em{\sc i\kern-.025em b}\kern-.08em
    T\kern-.1667em\lower.7ex\hbox{E}\kern-.125emX}}
\begin{document}

\title{Multimodal Deep Survival Analysis for Sinkhole Susceptibility}

\author{

\IEEEauthorblockN{Lucas Yuan}
\IEEEauthorblockA{\textit{Huron High School}\\
\textit{University of Florida}\\
Ann Arbor, MI, USA\\
lucasyuan1981@gmail.com}

\and

\IEEEauthorblockN{Minhee Kim}
\IEEEauthorblockA{\textit{Industrial and Systems Engineering}\\
\textit{University of Florida}\\
Gainesville, FL, USA\\
mkim3@ufl.edu}

\and

\IEEEauthorblockN{Zihan Li}
\IEEEauthorblockA{\textit{Industrial and Systems Engineering}\\
\textit{University of Florida}\\
Gainesville, FL, USA\\
zihanli1@ufl.edu}

\and

\IEEEauthorblockN{Chunli Dai}
\IEEEauthorblockA{\textit{Forest, Fisheries, and Geomatics Sciences}\\
\textit{University of Florida}\\
Gainesville, FL, USA\\
chunlidai@ufl.edu}

\and

\IEEEauthorblockN{Sanduni S. Disanayaka Mudiyanselage}
\IEEEauthorblockA{\textit{Forest, Fisheries, and Geomatics Sciences}\\
\textit{University of Florida}\\
Gainesville, FL, USA\\
sanduni.disanaya@ufl.edu}

\and

\IEEEauthorblockN{Ming Ye}
\IEEEauthorblockA{\textit{Department of Scientific Computing}\\
\textit{Florida State Univeristy}\\
Tallahassee, FL, USA\\
mye@fsu.edu}

\and

\IEEEauthorblockN{Kani Fu\IEEEauthorrefmark{1}}
\IEEEauthorblockA{\textit{Industrial and Systems Engineering}\\
\textit{University of Florida}\\
Gainesville, FL, USA\\
fu.kani@ufl.edu}
}

\maketitle

\begin{abstract}
Sinkholes are a widespread geohazard in karst terrain. In Florida, soluble carbonate bedrock, shallow groundwater, and intense rainfall combine to make subsidence both common
and spatially heterogeneous. Predicting where and when sinkholes will occur is difficult for two reasons. First, locations without reported sinkholes cannot be directly labeled or sampled as true negative locations. Second, the potential factors governing sinkhole risk span heterogeneous data modalities and therefore require careful integration within a unified modeling framework.
We address both problems with our proposed model, a multimodal Cox proportional hazards framework for sinkhole susceptibility. Our contributions are threefold. First, we extend the proportional-hazards formulation to heterogeneous multimodal input through modality-specific encoders and a cross-modal fusion layer. Second, we treat unreported locations as right-censored rather than negative, avoiding hard-negative labeling and yielding continuous, time-aware susceptibility from the predicted survival function. Third, a statewide Florida case study with spatially blocked validation and ablation studies quantifies the benefit of multimodal integration.
A Florida case study demonstrates that the proposed method effectively ranks sinkhole risk and produces a statewide susceptibility map that captures spatial variations in sinkhole occurrence.
\end{abstract}

\begin{IEEEkeywords}
Sinkhole susceptibility; Multimodal learning; Cox hazards model; Survival analysis.
\end{IEEEkeywords}

\vspace{-5pt}
\section{Introduction}

Sinkholes are a major geohazard in karst regions underlain by soluble rocks such as limestone and dolomite. In Florida, karst processes interact with groundwater fluctuations, precipitation, land development, and geological structure, producing substantial spatial variation in susceptibility. Historical subsidence reports document past failures but are imperfect labels: many reported depressions are not field-verified sinkholes, and many susceptible sites go unreported because of reporting bias, limited monitoring, or incomplete records.
A further challenge is that the factors governing sinkhole formation are inherently multimodal. Tabular geological databases encode site-level lithology, soil type, and
hydraulic attributes; digital elevation models and land-cover grids provide continuous
raster fields; and hydrological monitoring yields time series such as annual precipitation
and groundwater levels. These modalities differ in resolution, temporal cadence, and scale, and each captures a distinct facet of the environment.
To address these challenges, we develop an extended Cox proportional hazards model for multimodal sinkhole susceptibility assessment. The survival formulation represents locations without recorded incidents as right-censored rather than negative, indicating only that no event was observed before the censoring time. To handle heterogeneous covariates, we replace the linear log-hazard with a neural architecture of modality-specific encoders and a cross-modal fusion layer, preserving the spatial, temporal, and tabular structure of the inputs while letting their complementary information jointly determine the hazard.

Our contributions are as follows. (i) We extend the Cox model to heterogeneous multimodal geospatial data via modality-specific encoders and a cross-modal fusion layer, integrating tabular, raster, and time-series inputs without prematurely collapsing their spatial and temporal structure. (ii) We treat unreported locations as right-censored and estimate the model by Cox partial likelihood, avoiding hard-negative labeling and yielding continuous, time-aware susceptibility from the predicted survival function. (iii) We apply the framework
to statewide susceptibility assessment in Florida, using spatially blocked validation and ablation studies to quantify the benefit of multimodal integration over a single-modality baseline.

\section{Related Literature}
% -----------------------------------------------------------------------
\subsection{Sinkhole Susceptibility Mapping}

Early susceptibility mapping relied on expert knowledge and statistical index methods
such as the analytical hierarchy process \citep{taheri2015}. As geospatial data became
more available, data-driven statistical models were adopted, most commonly logistic
regression \citep{taheri2019}, and more recently machine learning models capable of capturing nonlinear covariate effects, including random forests and multilayer perceptrons \citep{muili2025}.
Despite this progress, two structural limitations persist. First, most studies formulate susceptibility as binary classification over observed sinkholes and sampled non-sinkhole locations \citep{subedi2019,muili2025}. Because unrecorded sites may simply be
unreported, owing to reporting bias
\citep{galve2009}, treating them as negatives introduces label inaccuracy.
Second, such approaches yield static categories or binary maps that collapse a continuum of risk into a few classes and omit the temporal dimension of occurrence, conveying where sinkholes are likely but not how risk evolves or the probability of an event within a given period. These limitations motivate a survival-analysis formulation, in which sites without recorded incidents are treated as censored rather than negative, preserving temporal information while accommodating incomplete records.

\subsection{Deep Survival Models}

Neural extensions of the Cox proportional hazards model \citep{cox1972} replace its linear predictor with a network while retaining the partial-likelihood objective. Continuous-time variants can be found in \citep{katzman2018deepsurv,kvamme2019time}, which progressively relax the proportional-hazards assumption by letting the risk function depend
jointly on covariates and time. Discrete-time formulations that partition the time axis and
estimate interval-specific hazards, some adding ranking losses or recurrent structure to
capture cross-interval dependencies, are developed in
\citep{lee2018deephit,ren2019deep}.
These methods learn nonlinear covariate effects from censored data but offer limited
guidance on integrating multimodal geospatial inputs \citep{wiegrebe2024deep}. We therefore
develop a multimodal survival framework that jointly processes tabular lithological
attributes, raster hydraulic-head patches, annual precipitation sequences, and site
coordinates. Following DeepSurv, the linear log-risk is replaced with a network. Unlike its
static covariate representation, precipitation histories are encoded recurrently so that the
estimated risk reflects antecedent hydrological conditions.

% -----------------------------------------------------------------------
\section{Problem Definition}
% -----------------------------------------------------------------------

Let $\mathcal{S}=\{1,\ldots,N\}$ be a set of sites, and let $s_i$ denote the spatial information for unit
$i$, including latitude and longitude.  The geospatial covariates for unit $i$ are
partitioned into three modality blocks, $X_i = \bigl(X_i^{\mathrm{tab}},\; X_i^{\mathrm{ras}},\; X_i^{\mathrm{ts}}\bigr),$
where $X_i^{\mathrm{tab}}$ collects tabular site attributes, $X_i^{\mathrm{ras}}$ represents
spatially continuous raster-derived features, and $X_i^{\mathrm{ts}}$ represents time-indexed
precipitation records.  Section~\ref{sec:dataset} describes the contents of each modality block in detail. The goal is to estimate the sinkhole-risk probability
\begin{equation}
  p_i = P(Y_i=1\mid X_i, s_i),
\end{equation}
where $Y_i$ denotes sinkhole occurrence within the next $K$ years.

% -----------------------------------------------------------------------
\section{Methodology}
\label{sec:method}
% -----------------------------------------------------------------------

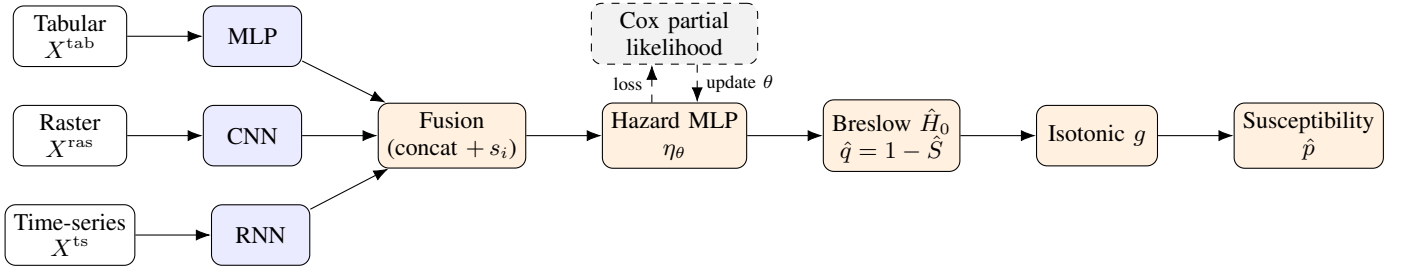
\begin{figure*}[t]
\centering
\begin{tikzpicture}[
  font=\small,
  box/.style={draw, rounded corners, minimum height=8mm, minimum width=15mm, align=center},
  enc/.style={draw, fill=blue!8, rounded corners, minimum height=8mm, minimum width=13mm, align=center},
  op/.style={draw, fill=orange!12, rounded corners, minimum height=8mm, minimum width=16mm, align=center},
  train/.style={draw, dashed, fill=gray!10, rounded corners, minimum height=8mm, minimum width=22mm, align=center},
  >={Latex[length=2mm]},
  node distance=5mm and 10mm
]
% inputs
\node[box] (tab) {Tabular\\$X^{\mathrm{tab}}$};
\node[box, below=of tab] (ras) {Raster\\$X^{\mathrm{ras}}$};
\node[box, below=of ras] (ts) {Time-series\\$X^{\mathrm{ts}}$};
% encoders
\node[enc, right=of tab] (etab) {MLP};
\node[enc, right=of ras] (eras) {CNN};
\node[enc, right=of ts]  (ets)  {RNN};
% main pipeline (left to right)
\node[op, right=of eras] (fuse) {Fusion\\(concat $+\,s_i$)};
\node[op, right=of fuse] (haz)  {Hazard MLP\\$\eta_\theta$};
\node[op, right=of haz]  (bres) {Breslow $\hat{H}_0$\\$\hat{q}=1-\hat{S}$};
\node[op, right=of bres] (cal)  {Isotonic $g$};
\node[op, right=of cal]  (out)  {Susceptibility\\$\hat{p}$};
% training objective (above the hazard MLP)
\node[train, above=of haz] (cox) {Cox partial\\likelihood};
% fan-in
\draw[->] (tab)--(etab); \draw[->] (ras)--(eras); \draw[->] (ts)--(ets);
\draw[->] (etab)--(fuse); \draw[->] (eras)--(fuse); \draw[->] (ets)--(fuse);
% main flow (solid = inference)
\draw[->] (fuse)--(haz);
\draw[->] (haz)--(bres);
\draw[->] (bres)--(cal);
\draw[->] (cal)--(out);
% training loop (dashed)
\draw[->,dashed] ([xshift=-3mm]haz.north) -- ([xshift=-3mm]cox.south)
  node[midway,left,font=\scriptsize]{loss};
\draw[->,dashed] ([xshift=3mm]cox.south) -- ([xshift=3mm]haz.north)
  node[midway,right,font=\scriptsize]{update $\theta$};
\end{tikzpicture}
\caption{Multimodal Cox survival framework.}
\label{fig:framework}
\end{figure*}

The framework fits a multimodal Cox proportional hazards model to the sample sites and produces survival probabilities at any spatial unit in the study area, which serve directly as susceptibility estimates.
Figure~\ref{fig:framework} summarizes the architecture.

\begin{algorithm}[t]
\caption{Multimodal DeepSurv}
\label{alg:mvp_deepsurv}
\footnotesize
\setlength{\algorithmicindent}{1em}
\begin{algorithmic}[1]
\Require Site-year data
modalities $\mathcal{M}=\{\mathrm{tab},\mathrm{ras},\mathrm{ts}\}$, horizon $K$
\Ensure Calibrated susceptibility $\{\hat{p}_i\}$

\Statex \textbf{Stage 1: Autoencoder pretraining}
\For{each modality $m\in\mathcal{M}$}
    \State Initialize encoder $g^m_{\theta_m}$ and decoder $\mathrm{Dec}^m_{\psi_m}$
    \Repeat
        \State $e_i^m \gets g^m_{\theta_m}(X_i^m)$
        \State $\hat{X}_i^m \gets \mathrm{Dec}^m_{\psi_m}(e_i^m)$
        \State Minimize $\mathcal{L}_{\mathrm{rec}}^m
        = \lVert X_i^m - \hat{X}_i^m \rVert_2^2$
    \Until{convergence}
    \State Retain encoder $g^m_{\theta_m}$; discard decoder $\mathrm{Dec}^m_{\psi_m}$
\EndFor

\Statex \textbf{Stage 2: Multimodal survival learning}
\Repeat
    \For{each site-year $(i,t)$}
        \State $e_i^m(t)\gets g^m_{\theta_m}\!\bigl(X_i^m(t)\bigr),\ m\in\mathcal{M}$
            \Comment{only $X_i^{\mathrm{ts}}$ varies with $t$}
        \State $e_i(t)\gets
        \mathrm{Fuse}_{\theta_4}\!\bigl(e_i^{\mathrm{tab}},e_i^{\mathrm{ras}},
        e_i^{\mathrm{ts}}(t),s_i\bigr)$
        \State $\eta_\theta(X_i,t)\gets \mathrm{MLP}_{\theta_5}\!\bigl(e_i(t)\bigr)$
    \EndFor
    \State Minimize the negative Cox partial likelihood (Breslow ties)
    \[
    \mathcal{L}(\theta)=-\sum_{k}
    \Bigl[\sum_{i\in\mathcal{D}_k}\eta_\theta(X_i,t_k)
    -d_k\log\!\!\sum_{j\in\mathcal{R}(t_k)}\!\!
    \exp\{\eta_\theta(X_j,t_k)\}\Bigr]
    \]
    \State Update $\theta=\{\theta_m\}_{m\in\mathcal{M}}\cup\{\theta_4,\theta_5\}$
\Until{convergence}

\Statex \textbf{Stage 3: Susceptibility prediction}
\State Estimate baseline cumulative hazard $\hat{H}_0(t)$ (Breslow)
\State Fit isotonic calibrator $g$ on validation windows
\For{each prediction site $i$}
    \State $\hat{S}_i(t+K\mid t)\gets
    \exp\!\bigl[-\{\hat{H}_0(t+K)-\hat{H}_0(t)\}
    \exp\{\eta_\theta(X_i,t)\}\bigr]$
    \State $\hat{q}_i\gets 1-\hat{S}_i(t+K\mid t)$
    \State $\hat{p}_i\gets g(\hat{q}_i)$
    \Comment{calibrated susceptibility}
\EndFor
\State \Return $\{\hat{p}_i\}$
\end{algorithmic}
\end{algorithm}
\vspace{-5pt}

\subsection{Survival Setup and Right-Censoring}
\label{sec:survival}

Each unit $i$ is a site that either has a reported subsidence or is right-censored, and we use calendar year $t$ as the timescale.  Let $T_i$ be the calendar year of the first reported subsidence at site $i$ and $C_i$ the final year of observation; if no event is reported through $C_i$, the site is right-censored.
The observed time and event indicator are
\begin{equation}
  \tilde{T}_i=\min(T_i,C_i),\qquad \Delta_i=\mathbb{I}(T_i\leq C_i),
\end{equation}
where $\Delta_i=1$ marks a reported subsidence and $\Delta_i=0$ marks right
censoring.  Censored sites are retained in the risk set up to $C_i$, contributing
the information that no event occurred by then rather than being treated as hard
negatives.  Because the model is fitted to start--stop intervals, each site-year row
$(i,\,t_{\mathrm{start}},\,t_{\mathrm{stop}},\,\Delta_{it},\,X_i^{\mathrm{ts}}(t))$
covers one calendar year, allowing time-series covariates to vary across rows while
the tabular and raster blocks remain fixed.

\subsection{Modality-Specific Encoders}
\label{sec:encoders}

Each modality block is processed by an encoder.

\paragraph{Tabular encoder.}
Categorical attributes are one-hot encoded and concatenated with continuous features to form the input $X_i^{\mathrm{tab}}$. A multilayer perceptron (MLP) maps this vector to a $d$-dimensional embedding:
\begin{equation}
  e_i^{\mathrm{tab}} = g^{\mathrm{tab}}_{\theta_1}(X_i^{\mathrm{tab}}) \in \mathbb{R}^d.
\end{equation}

\paragraph{Raster encoder.}
For each site $i$, a spatial patch of raster layers is extracted around location $s_i$.  A spatial encoder (CNN) maps the patch to an embedding:
\begin{equation}
  e_i^{\mathrm{ras}} = g^{\mathrm{ras}}_{\theta_2}(X_i^{\mathrm{ras}}) \in \mathbb{R}^d.
\end{equation}

\paragraph{Time-series encoder.}
Annual precipitation summaries form a multivariate sequence $X_i^{\mathrm{ts}} = (x_{i,t_0}, x_{i,t_0+1}, \ldots)$ up to year $t$.  A recurrent network (RNN) encodes the time-series:
\begin{equation}
  e_i^{\mathrm{ts}}(t) = g^{\mathrm{ts}}_{\theta_3}(X_{i,t}^{\mathrm{ts}}) \in \mathbb{R}^d.
\end{equation}

The three modality embeddings and the spatial context $s_i$ are combined by a fusion
module,
\begin{equation}
  e_i(t)
  = \mathrm{Fuse}\!\left(e_i^{\mathrm{tab}},\; e_i^{\mathrm{ras}},\;
    e_i^{\mathrm{ts}}(t),\; s_i\right) \in \mathbb{R}^{d_f},
\end{equation}
where $\mathrm{Fuse}$ is implemented as concatenation followed by a linear projection. A final MLP maps the fused representation to a log-relative hazard:
\begin{equation}
  \eta_\theta(X_i, t) = \mathrm{MLP}_{\theta_5}(e_i(t)) \in \mathbb{R},
\end{equation}
where $\theta = \{\theta_1, \theta_2, \theta_3, \theta_4, \theta_5\}$ collects all
encoder and fusion parameters. Then the full hazard model is represented as:
\begin{equation}
  h_i(t \mid X_i)
  = h_0(t)\exp\!\left\{\eta_\theta(X_i, t)\right\}.
\end{equation}

\subsection{Training via Partial Likelihood}
\label{sec:training}

All parameters $\theta$ are estimated by minimizing the negative Cox partial
likelihood. The partial likelihood loss is:
\begin{equation}
  \mathcal{L}(\theta)
  = -\sum_{i:\,\Delta_i=1}
    \left[
      \eta_\theta(X_i, \tilde{T}_i)
      - \log\sum_{j\in\mathcal{R}(\tilde{T}_i)}
        \exp\!\left\{\eta_\theta(X_j, \tilde{T}_i)\right\}
    \right],
\end{equation}
where $\mathcal{R}(t) = \{j:\tilde{T}_j\geq t\}$ is the risk set at time $t$.  Only
sites with observed events ($\Delta_i=1$) contribute to the sum over events.

\subsection{Susceptibility Estimation}
\label{sec:prediction}

After training, the baseline cumulative hazard is estimated from the training sites
via the Breslow estimator,
\begin{equation}
  \hat{H}_0(t)
  = \sum_{t_k\leq t}
    \frac{d_k}{\displaystyle\sum_{j\in\mathcal{R}(t_k)}
      \exp\!\left\{\eta_\theta(X_j, t_k)\right\}},
\end{equation}
where $d_k$ is the number of observed events at $t_k$.  For any spatial unit $i$, the model's survival probability over a $K$-year horizon starting from year $t$ is
\begin{equation}
  \hat{S}_i(t+K\mid t)
  = \exp\!\left[
      -\bigl\{\hat{H}_0(t+K) - \hat{H}_0(t)\bigr\}
      \exp\!\left\{\eta_\theta(X_i, t)\right\}
    \right],
\end{equation}
which yields a raw complementary incidence probability
\begin{equation}
\hat{q}_i = 1 - \hat{S}_i(t+K\mid t).
\end{equation}
Because subsidence events are rare, this probability is poorly calibrated in absolute terms: the network $\eta_\theta$ orders sites well, but the raw magnitudes
$\hat{q}_i$ do not match observed event frequencies. We therefore compose the model output with a monotone calibration map estimated using isotonic regression: $\hat{p}_i=g(\hat{q}_i)\in[0,1],$
where $g:[0,1]\to[0,1]$ is a non-decreasing step function obtained by minimizing $\sum_j \left(g(\hat{q}_j)-y_j\right)^2$ 
over the validation windows. Here, $\hat{q}_j$ denotes the uncalibrated prediction for validation window $j$, and $y_j\in{0,1}$ is its observed binary outcome, with $y_j=1$ indicating that a sinkhole event is observed and $y_j=0$ otherwise. The optimization is solved using the pool-adjacent-violators algorithm in validation set.

\section{Dataset and Preprocessing}
\label{sec:dataset}

\begin{figure}[htbp]
\centering
\includegraphics[width=0.6\columnwidth]{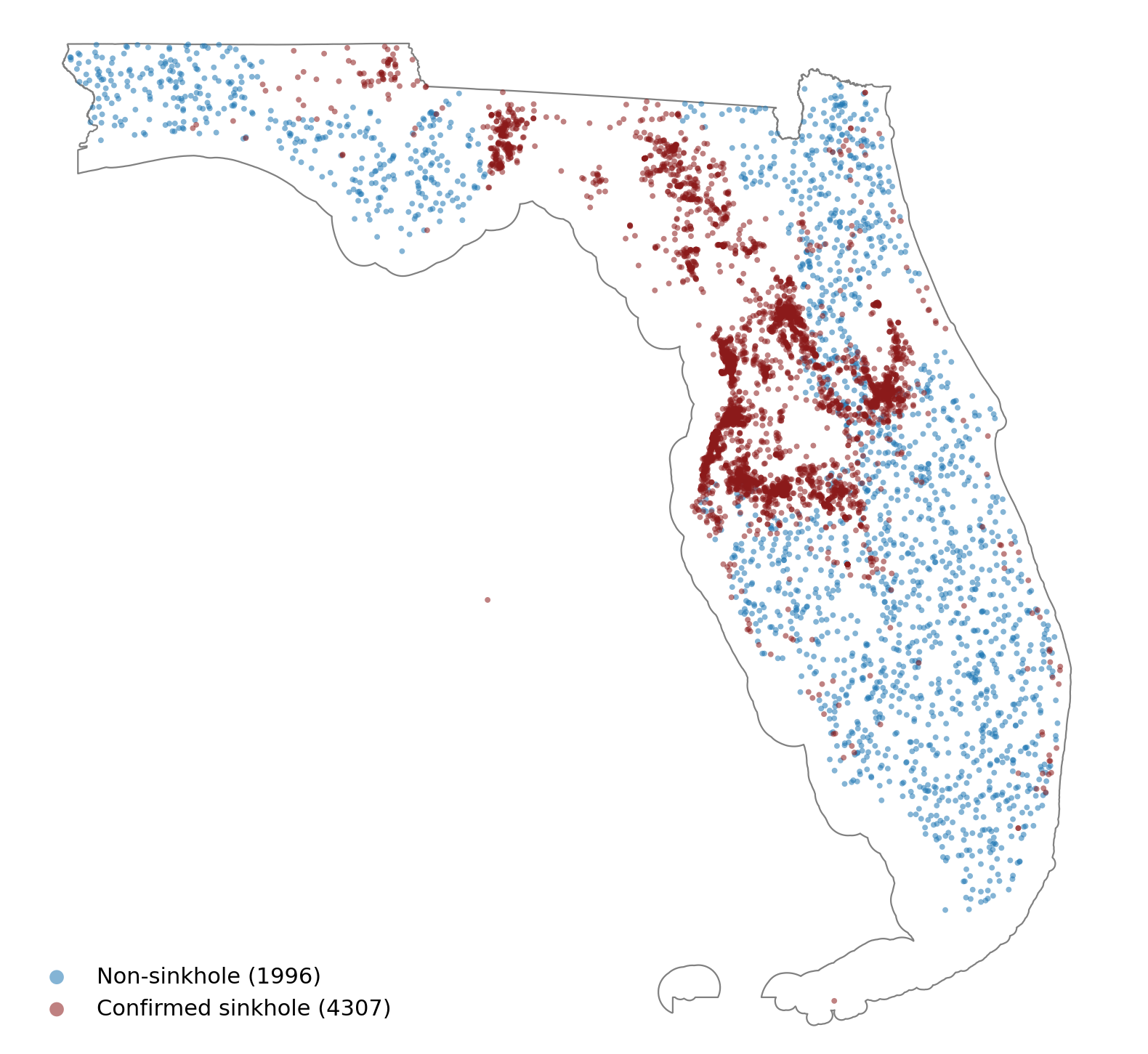}
\caption{Spatial distribution of sinkhole and non-sinkhole sites.}
\label{fig:all-sites}
\end{figure}

The study covers Florida and combines three public datasets: subsidence incident locations from the Florida Geological Survey,\footnote{\url{https://floridadep.gov/fgs/sinkholes/content/subsidence-incident-reports}} annual precipitation from PRISM for 1895--2025,\footnote{\url{https://prism.oregonstate.edu/}} and the hydraulic-head difference between the surficial water table and the Floridan Aquifer System.\footnote{\url{https://geodata.dep.state.fl.us/search?tags=head\%2520difference}} After screening the subsidence records, 6{,}302 sinkhole and non-sinkhole sites were retained (Figure~\ref{fig:all-sites}).

Each site was converted into annual sliding samples using a 10-year precipitation observation window followed by a 10-year prediction horizon. For each sinkhole site, only the window ending in its recorded event year was labeled as an event; all other windows were censored. This process produced 557{,}946 samples with a low event rate.

\begin{figure}[htbp]
\vspace{-6pt}
\centering
\includegraphics[width=1.0\columnwidth]{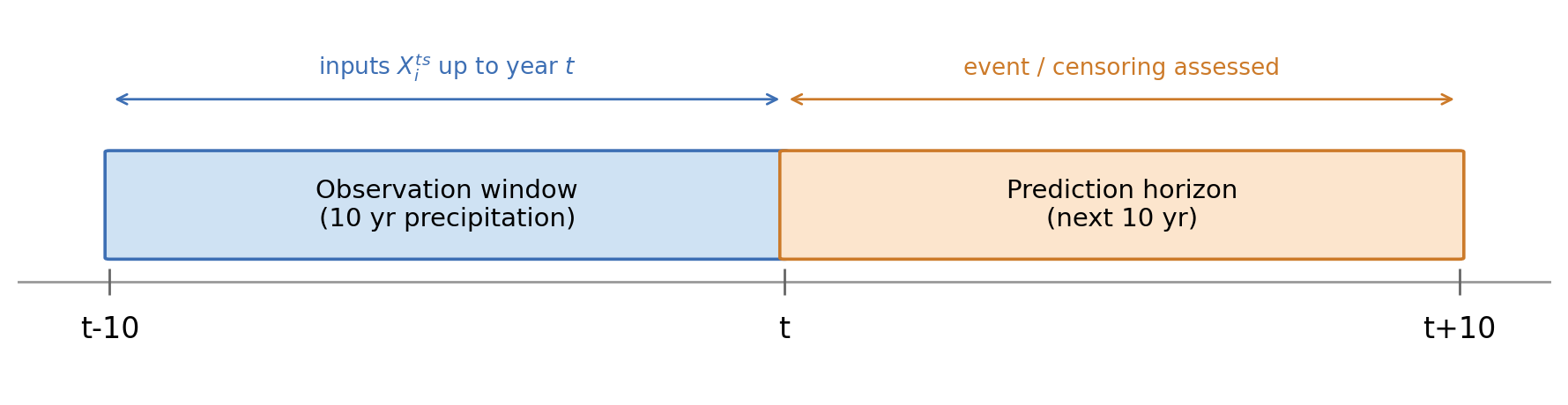}
\vspace{-6pt}
\caption{Sliding-window construction with a 10-year observation window and a 10-year prediction horizon.}
\label{fig:window}
\end{figure}

\vspace{-5pt}
\section{Results and Discussion}

\subsection{Evaluation Setup}
\label{sec:evaluation-setup}

We report three metrics on the held-out test windows. Let $\hat\eta_i$ denote the model's
risk score (log-hazard) for window $i$, $\hat p_i$ its calibrated 10-year probability,
$(t_i,\delta_i)$ the observed time and event indicator, and $y_i\in\{0,1\}$ the binary
event label. Probabilities are calibrated by the isotonic map fit on the validation windows
and applied unchanged to the test windows (Section~\ref{sec:prediction}).

\paragraph{Concordance index (C-index).} The C-index measures ranking accuracy under
censoring: the fraction of comparable pairs whose predicted risk ordering agrees with the
observed event-time ordering. A pair $(i,j)$ is comparable when the earlier-timed subject
experienced an event ($\delta_i=1,\; t_i<t_j$), and concordant when the model assigns it the
higher risk:
\begin{equation}
  \mathrm{C\text{-}index}=
  \frac{\displaystyle\sum_{i,j}\delta_i\,\mathbf{1}[t_i<t_j]\,\mathbf{1}[\hat\eta_i>\hat\eta_j]}
       {\displaystyle\sum_{i,j}\delta_i\,\mathbf{1}[t_i<t_j]}\,,
  \label{eq:cindex}
\end{equation}
with $0.5$ corresponding to random ordering and $1.0$ to perfect ordering.

\paragraph{Windowed ROC-AUC.} Treating each window as event/non-event, the AUC is the
probability that a randomly chosen event window is scored above a randomly chosen non-event
window. With $\mathcal{P}$ and $\mathcal{N}$ the event and non-event sets,
\begin{equation}
  \mathrm{AUC}=
  \frac{1}{|\mathcal{P}|\,|\mathcal{N}|}
  \sum_{i\in\mathcal{P}}\sum_{j\in\mathcal{N}}
  \Big(\mathbf{1}[\hat p_i>\hat p_j]+\tfrac{1}{2}\,\mathbf{1}[\hat p_i=\hat p_j]\Big).
  \label{eq:auc}
\end{equation}
Unlike the C-index, the AUC ignores event timing and censoring and measures binary
discrimination only.

\paragraph{Expected calibration error (ECE).} Partitioning predictions into $M$ equal-width
probability bins $\{B_m\}$, ECE is the sample-weighted gap between each bin's mean predicted
probability and its empirical event frequency:
\begin{equation}
  \mathrm{ECE}=\sum_{m=1}^{M}\frac{|B_m|}{n}\,
  \big|\,\mathrm{acc}(B_m)-\mathrm{conf}(B_m)\,\big|,
  \label{eq:ece}
\end{equation}
where $\mathrm{conf}(B_m)=\tfrac{1}{|B_m|}\sum_{i\in B_m}\hat p_i$,
$\mathrm{acc}(B_m)=\tfrac{1}{|B_m|}\sum_{i\in B_m}y_i$, and $n$ is the total sample count.

\vspace{4pt}
We evaluate under two holdout regimes. The \textbf{random split} assigns individual sliding
windows to train/validation/test uniformly at random, measuring interpolation; because a
site's windows overlap heavily in their 10-year inputs, nearby windows of one site can appear
in both train and test. The \textbf{county (geologic-unit) split} assigns whole geologic units
to a single fold, so held-out sites are spatially disjoint from training, measuring
generalization to unseen terrain under a fixed split seed.

\subsection{Results}
\label{sec:main-results}

\begin{table}[htbp]
    \centering
    \caption{Test performance under both split regimes. \emph{Lift} is the mean calibrated susceptibility assigned to confirmed sinkholes withheld from training.}
    \label{tab:multimodal_deepsurv_comparison}
    \renewcommand{\arraystretch}{1.15}
    \begin{tabular}{@{}llcccc@{}}
        \toprule
        \textbf{Split} & \textbf{Method} & \textbf{C-index} & \textbf{ROC-AUC}
        & \textbf{ECE} & \textbf{Lift} \\
        \midrule
        Random & Time-series only     & 0.743 & 0.879 & 0.0007 & $4.4\times$ \\
        Random & Multimodal           & 0.780 & 0.899 & 0.0008 & $7.1\times$ \\
        Random & \textbf{+\,pretrain} & \textbf{0.780} & \textbf{0.899} & 0.0007 & $7.0\times$ \\
        \midrule
        County & Time-series only     & 0.696 & 0.888 & 0.0021 & $6.4\times$ \\
        County & Multimodal           & 0.695 & 0.904 & 0.0011 & $6.9\times$ \\
        County & \textbf{+\,pretrain} & \textbf{0.703} & 0.899 & 0.0021 & $6.8\times$ \\
        \bottomrule
    \end{tabular}
\end{table}

\begin{figure}[htbp]
  \centering
  \includegraphics[width=0.9\columnwidth]{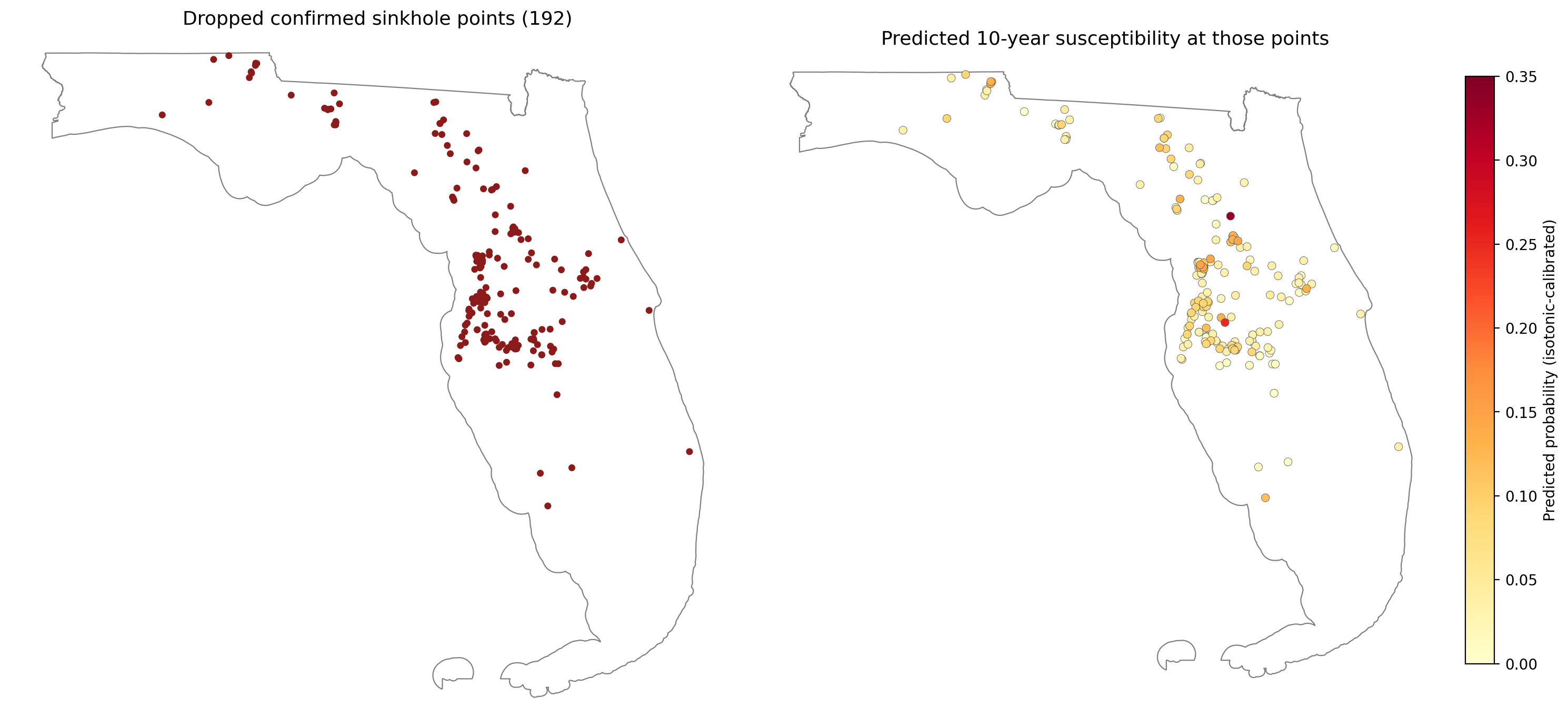}
  \caption{Point validation. Left: 192 confirmed testing sinkholes. Right: the same points resulting from the proposed algorithm on average $7\times$ background.}
  \label{fig:sinkhole-val-maps}
\end{figure}

\begin{figure}[t]
    \centering
    \includegraphics[width=0.55\columnwidth]{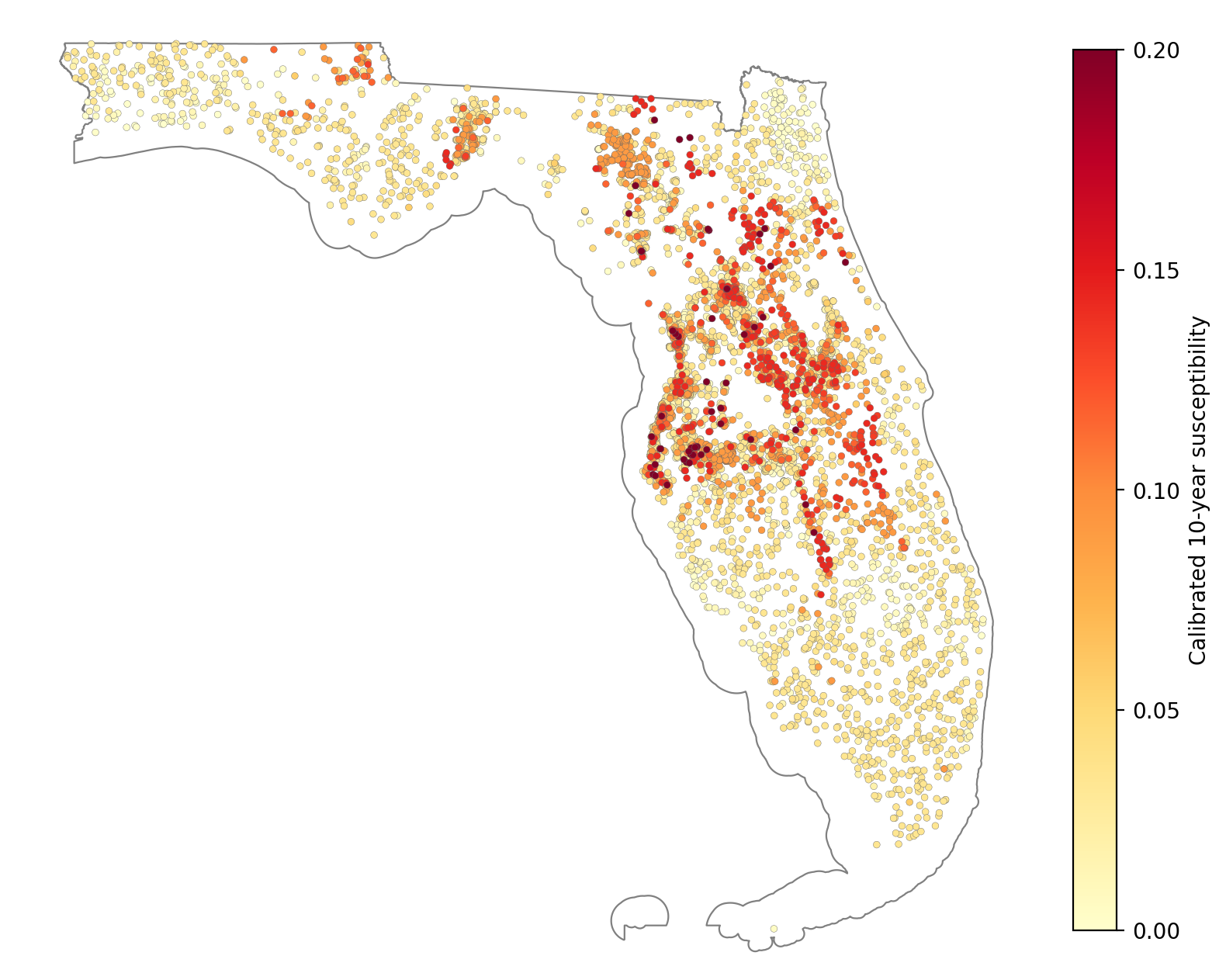}
    \caption{Statewide calibrated 10-year sinkhole susceptibility at all monitoring sites, each
    scored on its most recent observed window. High susceptibility concentrates in the
    west-central karst belt, coincident with the confirmed-sinkhole distribution of
    Figure~\ref{fig:all-sites}.}
    \label{fig:susceptibility-map}
\end{figure}

Table~\ref{tab:multimodal_deepsurv_comparison} reports every configuration under both splits.
The pretrained multimodal model reaches C-index 0.780 on the random split versus 0.703 on the
county split; the 0.077 gap reflects interpolation between near-duplicate windows of a site
(adjacent windows share 9 of 10 observation years), so the county number is the more defensible
estimate for unseen terrain, while ROC-AUC is identical across regimes (0.899), locating the
gap in censoring-aware time-concordance rather than binary discrimination. Fusing the static
modalities with precipitation is the primary driver of performance, improving both C-index and
ROC-AUC and raising the calibrated risk assigned to confirmed sinkholes withheld entirely from
training to roughly $7\times$ background (Figure~\ref{fig:sinkhole-val-maps}); that these unseen
sinkholes still score high, and concentrate in the west-central karst belt rather than being
uniformly raised, indicates the model captures genuine susceptibility rather than memorized
locations, consistent with subsidence being controlled by subsurface lithology that
precipitation alone cannot localize. Encoder pretraining yields no consistent gain, as
fine-tuning recovers equivalent encoders regardless of initialization. The raw Cox
probabilities rank well but are miscalibrated; a monotone post-hoc isotonic step removes the
calibration error at negligible cost to discrimination, letting the outputs be read as absolute
10-year probabilities. Scored on each site's most recent decade, the resulting statewide map
(Figure~\ref{fig:susceptibility-map}) concentrates high susceptibility along the west-central
karst belt, coincident with the confirmed-sinkhole distribution without being fit to those
labels.

\vspace{-10pt}

\section{Conclusion}

We presented a multimodal deep survival framework for sinkhole susceptibility across
Florida, integrating a precipitation time series with static geospatial modalities and
evaluated under both interpolation-oriented and spatially disjoint partitions. Both the
multimodal representation and the survival formulation contribute to predictive
performance, with reliable discrimination under the spatially disjoint partition---the
more faithful measure of generalization to unseen terrain. Beyond discrimination, post-hoc
isotonic recalibration is the most robust contribution, giving the predicted
susceptibilities absolute probabilistic meaning while preserving ranking quality. Future
work includes characterizing county-split variance across repeated held-out-unit draws and
extending the calibrated map into a decision-support tool for karst-prone regions.

\vspace{-2pt}
\section*{Acknowledgment}
This work was supported by the National Science Foundation under Grant No. 2530726.

\vspace{-1pt}
\bibliographystyle{IEEEtranN}
\bibliography{references}

\end{document}